%% file: ms.tex
\pdfoutput=1
\documentclass{article}

\usepackage[preprint]{neurips_2023}

\usepackage[utf8]{inputenc} 
\usepackage[T1]{fontenc}    
\usepackage{natbib}
\usepackage{hyperref}       
\usepackage{url}            
\usepackage{booktabs}       
\usepackage{amsfonts}       
\usepackage{nicefrac}       
\usepackage{microtype}      
\usepackage{xcolor}         
\usepackage{svg}
\usepackage{amsmath}
\usepackage{bm}
\usepackage{amssymb}
\usepackage{pifont}
\usepackage{multirow}

\newcommand{\model}{PointCAM}
\newcommand{\lmpm}{\mathcal{L}_\text{MPM}}
\newcommand{\lcls}{\mathcal{L}_\text{CLS}}

\title{Self-supervised adversarial masking for 3D point cloud representation learning}

\author{%
  Michał Szachniewicz\normalfont{\textsuperscript{1}}, \textbf{Wojciech Kozłowski}\normalfont{\textsuperscript{1}}, \textbf{Michał Stypułkowski}\normalfont{\textsuperscript{2}}, \textbf{Maciej Zięba}\normalfont{\textsuperscript{1,3}}
  \AND
  \normalfont{\textsuperscript{1}}Wrocław University of Science and Technology
  \And \normalfont{\textsuperscript{2}}University of Wrocław
  \And \normalfont{\textsuperscript{3}}Tooploox
}

\begin{document}

\maketitle

\begin{abstract}
  Self-supervised methods have been proven effective for learning deep representations of 3D point cloud data. Although recent methods in this domain often rely on random masking of inputs, the results of this approach can be improved. We introduce \model\, a novel adversarial method for learning a masking function for point clouds. Our model utilizes a self-distillation framework with an online tokenizer for 3D point clouds. Compared to previous techniques that optimize patch-level and object-level objectives, we postulate applying an auxiliary network that learns how to select masks instead of choosing them randomly. Our results show that the learned masking function achieves state-of-the-art or competitive performance on various downstream tasks. The source code is available at \url{https://github.com/szacho/pointcam}.
\end{abstract}

\input{sections/0_introduction}
\input{sections/1_related_work}

\input{sections/2_method}

\input{sections/3_experiments}

\input{sections/4_conclusion}

\bibliographystyle{plain}
\bibliography{ms}

\end{document}

%% file: sections/0_introduction.tex
\section{Introduction}

Deep learning has revolutionized various domains of computer vision, including the processing and analysis of 3D point cloud data. Learning effective representations from point clouds is a fundamental task for many applications such as autonomous driving, robotics, and augmented reality. Self-supervised learning methods have emerged as a promising approach for unsupervised representation learning, where models are trained to learn meaningful features without the need for explicit human annotations.

In recent years, self-supervised methods have shown success in learning representations of various data modalities, including images, text, and speech, \cite{chen2020simple, van2020scan, he2020momentum, caron2020unsupervised, caron2021emerging, zhou2022ibot, devlin2019bert, baevski2022data2vec, chorowski2021aligned} enabling better performance on downstream tasks with limited annotated data points. Methods created for 3D point clouds \cite{yu2022pointbert, liu2022masked, pang2022masked, zhang2022pointm2ae} typically employ random masking of input points, where the model is trained to predict the missing parts of the object. 

However, the goal of masking is to make it harder for the model to recover from corrupted views with a meaningful representation. Random masks are therefore not an ideal approach to the problem. They are not capable of covering the whole semantic parts of a given object, usually keeping most information from an input untouched.

To address these limitations, we propose \model, a novel adversarial method for learning a masking function that employs a self-distillation framework with an online tokenizer specifically designed for 3D point clouds. Instead of using randomly generated masks, we introduce an auxiliary network that learns to select masks in a trainable manner. By doing so, we aim to optimize the masking function, enabling it to effectively capture relevant contextual information and promote more informative feature learning. The model is trained in a student-teacher manner, where the former aims to maximize the cross-entropy loss between the projected mask token representation and the corresponding projected patch representation from the teacher.

We investigate architecture improvements and are the first to adapt the state-of-the-art image-based framework iBOT \cite{zhou2022ibot} to the point cloud domain, enhancing the performance of our model. This also allows us to train an online tokenizer jointly with MPM objective without the need of using a pre-trained tokenizer as in PointBERT \cite{yu2022pointbert} and thus making our framework trainable in an end-to-end manner. 

Using our proposed method, we train a transformer encoder and demonstrate its effectiveness in achieving state-of-the-art performance within the scope of methods based on non-hierarchical transformer \cite{vaswani2017attention} architectures for most of the standard classification and segmentation downstream tasks. While our results are comparable or superior within this category, it should be noted that more sophisticated architectures, specifically hierarchical transformers \cite{zhang2022pointm2ae}, achieve higher performance. However, our findings highlight the potential of our approach in maximizing the capabilities of standard transformer models and provide a foundation for future work to bridge the performance gap with more complex architectures.

The main contributions of this paper can be summarized as follows:
    \textbf{(i)} We replace the random masking function in the patch-level objective with a trainable masking function, allowing it to learn how to corrupt semantically meaningful parts of point clouds.
    \textbf{(ii)} To the best of our knowledge, we are the first to adapt the state-of-the-art image self-supervised framework iBOT \cite{zhou2022ibot} to the point cloud domain, enabling end-to-end optimization of all of the modules.
    \textbf{(iii)} We evaluate \model\ on classification and segmentation downstream tasks, achieving results superior to the current state-of-the-art non-hierarchical techniques on most of them.

%% file: sections/1_related_work.tex
\section{Related work}

\paragraph{Self-supervised learning for images} The self-supervised learning is extensively studied for learning image representations \cite{chen2020simple, van2020scan, he2020momentum, caron2020unsupervised}. Most recently, the transformer-based approaches gained excellent quality in learning representations. Dino \cite{caron2021emerging} utilizes vision transformers to train high-quality representations. iBOT \cite{zhou2022ibot} enriches Dino's concepts by masked prediction with an online tokenizer. ADIOS \cite{shi2022adversarial} is another masking framework for self-supervised learning, which simultaneously learns a masking function and an image encoder using an adversarial objective. Masked Autoencoders (MAE) \cite{he2022masked} are also shown to be scalable self-supervised learners for computer vision.

\paragraph{Learning representations for point clouds} The problem of representation learning is also widely investigated for point clouds. Some approaches utilize the Generative Adversarial Networks (GANs) concepts to train good-quality representations \cite{han2018view, wu2017learning}. Other groups of models construct the representations from autoencoders \cite{yang2018foldingnet, achlioptas2018learning}, and generative models like, variational autoencoders (VAEs) \cite{han2019multiangle} or adversarial autoencoders \cite{zamorski2020adversarial}. The authors of \cite{li2018sonet} postulate using Self-Organizing Network to extract informative features about the 3D objects. In \cite{valsesia2018learning} graph convolutions are used to construct discriminative representations.

\paragraph{Self-supervised learning for 3D point clouds} Recent models utilize self-supervised techniques to create meaningful representations for 3D point clouds. The authors of \cite{zhang2021selfsupervised} use a simple self-supervised pre-training method that can work with single-view depth scans without 3D registration and point correspondences. PointContrast \cite{xie2020pointcontrast} uses contrastive training as a pre-training strategy for downstream tasks. In \cite{sauder2019selfsupervised} neural network is trained to reconstruct point clouds whose parts have been randomly rearranged. In paper \cite{yan2023implicit} the authors propose to use an implicit decoder that reconstructs a continuous representation of the 3D shape that is independent of the imperfections in the discrete samples.
Most recent approaches adapt masking techniques known from the image domain. PointBERT \cite{yu2022pointbert} uses transformers with masked point modeling. Point-MAE \cite{pang2022masked} adapts the concept of masked autoencoders to point clouds. For 3D-OAE \cite{zhou20223doae}, masking is performed by occluding some local patches of point clouds and establishing the supervision via inpainting the occluded patches using the remaining ones. The pretraining technique from \cite{wang2021unsupervised} applies occlusions for various camera views and trains the encoder-decoder model to reconstruct the occluded points. MaskPoint \cite{liu2022masked} introduces a discriminative mask pretraining Transformer framework. CrossPoint \cite{afham2022crosspoint} takes advantage of multimodality, enabling a 3D-2D correspondence of objects by maximizing agreement between point clouds and the corresponding rendered 2D image. Hierarchical methods play a dominant role among other approaches considering the quality of representations evaluated on downstream tasks. Point-M2AE \cite{zhang2022pointm2ae} incorporates the multiscale masking and modified pyramid architecture for the encoder and decoder to progressively model spatial geometries and capture both fine-grained and high-level semantics of 3D shapes. 

Compared to existing approaches, our model first introduces trainable adversarial masking for 3D point clouds. Moreover, we adapt the iBOT architecture designed for images to the field of learning representations for 3D shapes.

%% file: sections/2_method.tex
\section{\model{}}
\begin{figure}
    \centering
    \includegraphics[width=\linewidth]{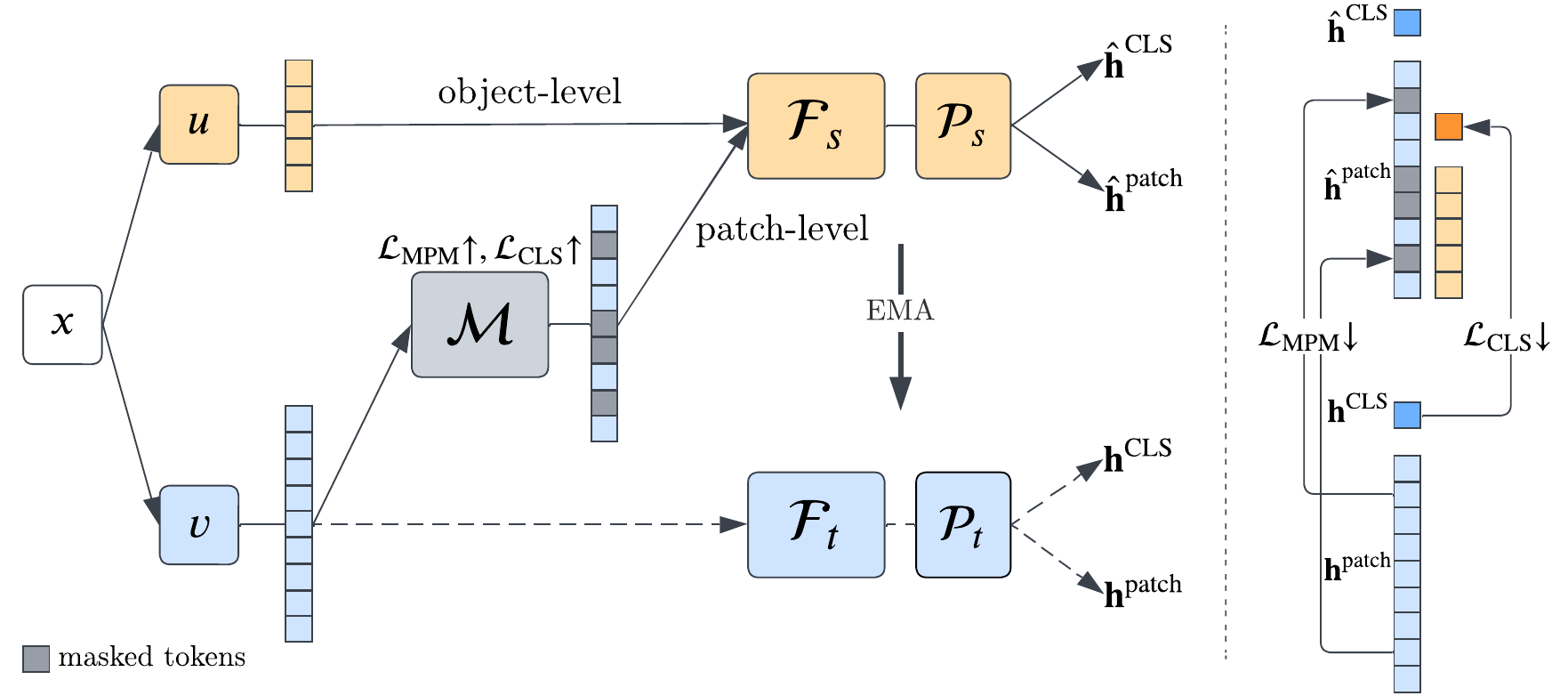}
    \caption{\textbf{Overview of \model\ framework.} Given a point cloud $x$, two views $u, v$ are created and divided into patches and processed with a simple PointNet \cite{qi2017pointnet}. The feature vector of view $u$ is passed to the student network $\mathcal{P}_s \circ \mathcal{F}_s$, while the feature vector of view $v$ proceeds to the mask generator $\mathcal{M}$. Its purpose is to mask patches in a way that maximizes both $\lmpm$ and $\lcls$ losses.
        Masked features are then recovered by $\mathcal{P}_s \circ \mathcal{F}_s$ to get projected features that match the ones predicted by the teacher network $\mathcal{P}_t \circ \mathcal{F}_t$.
        Object-level objective $\lcls$ is optimized similarly but on features from the CLS token.
        Weights of the teacher network are updated using exponential moving averaging of the student's parameters.
        After pre-training, we discard the projector $\mathcal{P}_s$ and derive the final representation of a point cloud from $\mathcal{F}_s$. The dashed arrows of the teacher branch symbolize a stop-gradient operator.}
    \label{img:pointcam}
\end{figure}

We design a self-supervised framework for learning deep representations of \textbf{Point} \textbf{C}louds with \textbf{A}dversarial \textbf{M}asking, abbreviated \textbf{\model{}}. An overview of the framework is presented in Figure \ref{img:pointcam}. Input in the form of a point cloud is grouped into patches (Section \ref{sec:patch_selection}) and two distinct views are created. These views are then passed to the teacher-student framework with a transformer backbone (Section \ref{sec:transformer}). In addition, one view is given to a mask generator (Section \ref{sec:mask_generator}) that determines which patches to mask, as opposed to selecting them randomly, making it a novel approach to masking in point cloud representation learning. The model is trained in a self-supervised manner (Section \ref{sec:ssl}), where the masked patches are recovered by the student network in order to satisfy the patch-level objective. For object-level objective we use features derived from the CLS token of the transformer. We employ cross-entropy loss as a similarity measure for both objectives and update teacher's parameters using the exponential moving average of student's weight, as in \cite{zhou2022ibot, caron2021emerging}.

\subsection{Input construction}
\label{sec:patch_selection}
To be able to process a point cloud with a transformer model, an input sequence of tokens has to be created. It is a straightforward process in text and vision domains, however, point clouds require a more sophisticated approach. Thus, we divide a point cloud into $p$ patches by choosing $p$ points using furthest point sampling (FPS) as center points and gather $k$ nearest points to each center in the Euclidean space.
This approach has a low computational cost and almost full coverage of the point set (about $99\%$) that we found sufficient for the training. These patches are then projected to patch embeddings with a small PointNet \cite{qi2017pointnet} that consists of 2 convolution blocks and max pooling operation, creating a patch embedding sequence $\{\mathbf{x}_i\}_{i=1}^p$. Center points are projected to position embeddings by applying a multilayer perceptron (MLP) comprising 2 linear layers and GELU \cite{hendrycks2020gaussian} activation function after the first layer, creating a position embedding sequence $\{\mathbf{x}^\text{POS}_i\}_{i=1}^p$. By adding patch and position embeddings, we obtain a token sequence $\mathbf{x} = \{\mathbf{x}_i + \mathbf{x}^\text{POS}_i\}_{i=1}^p$ that can be processed by a transformer architecture. We also add a CLS token, a pooling scheme introduced in BERT \cite{devlin2019bert}, at the first position of the token sequence.

\subsection{Transformer-based encoder}
\label{sec:transformer}
We use a transformer \cite{vaswani2017attention} with SwiGLU feed-forward layer, following the promising results of \cite{shazeer2020glu}, as a backbone that processes a token sequence $\mathbf{x}$ with each of transformer blocks. The final representation of a point cloud is obtained by concatenating the CLS token features with a sum of patch features aggregated using max and global average pooling operators.

\subsection{Mask generator}
\label{sec:mask_generator}
The main objective of the mask generator network $\mathcal{M}$ is to generate masks that occlude meaningful parts of a point cloud. It adds an adversarial aspect to the training, aiming to maximize losses of self-supervised tasks in contrast to the transformer's goal to minimize them. Intuitively, the mask generator should produce masks that make it harder for the student network to recover.
Network $\mathcal{M}$ follows a similar architectural design as the backbone, however, it is significantly smaller and there is no CLS token added. An additional MLP head is used to obtain continuous masks for a point cloud of size $N \times p$. Then, softmax activation is applied over the first dimension resulting in $N$ complementary masks for each of $p$ patches that are represented as a matrix $\mathbf{M} = (m_{ij}) = \mathcal{M}(\mathbf{x})$. The embedding of $j$-th patch is masked by $i$-th mask by multiplying its features by $1 - m_{ij}$, where $m_{ij} \in (0,1)$.

This construction does not have any restriction on covering all of the patches with one of the $N$ masks, making the prediction impossible for the student. To prevent generated masks from occluding every patch we follow previous work by adding a sparsity penalty \cite{shi2022adversarial}:
\begin{equation}
    \mathcal{L}_{\text{spar}}= \frac{1}{N} \sum_{i=1}^N \sin{\left( \frac{\pi}{p}  \sum_{j=1}^p m_{ij} \right)}^{-1}.
\end{equation}
where $m_{ij}$ corresponds to $j$-th patch in $i$-th mask. We experimentally found that masks with values close to $\{0, 1\}$ worked better than soft masks, and to encourage the network to produce such masks we add a diversity constraint \cite{zheng2019learning}:
\begin{equation}
    \mathcal{L}_{\text{div}} = \frac{1}{N^2}  \sum_{i=1}^N\sum_{k=1}^N  \left(1 - \frac{\mathbf{m}_{i} \mathbf{m}_{k}}{ \|\mathbf{m}_{i} \|_2 \| \mathbf{m}_{k} \|_2} \right).
\end{equation}

By optimizing these two objectives simultaneously we are able to generate masks that are mutually exclusive, close to binary, and almost evenly distributed in the point cloud.

\subsection{Self-supervised learning framework}
\label{sec:ssl}
We create two distinct views, $u$ and $v$, of a point cloud by cropping and augmenting it (for details see Section \ref{sec:exp_ssl}). Typically, the $v$ view covers a larger part of the point cloud than the $u$, allowing them to be referred to as global and local views.

Through the self-distillation process leveraging exponential moving average (EMA), we aim to learn representations that will be indifferent to these views. We create a token sequence $\mathbf{x}^u$, which we then pass to the student network obtaining projected features from the CLS and patch tokens $\{\mathbf{\hat{h}}^{\text{CLS}},\ \mathbf{\hat{h}}^\text{patch}\} = \mathcal{P}_s( \mathcal{F}_s( \mathbf{x}^u ))$. Similarly, we extract features from the teacher network $\{\mathbf{h}^{\text{CLS}},\ \mathbf{h}^\text{patch}\} = \mathcal{P}_t( \mathcal{F}_t( \mathbf{x}^v ))$. Additionally, we generate masks $\mathbf{M}$ and apply each one to $\mathbf{x}^v$, creating an input sequence that is passed to the student network:
\begin{equation}
    \label{eq:masking}
    \mathbf{\hat{x}}^{(i)} = (1 - \mathbf{m}_{i}) \mathbf{x}^v + \mathbf{m}_{i} \mathbf{x}^{\text{MASK}},
\end{equation}
where $\mathbf{x}^{\text{MASK}}$ is a trainable mask token, and $i = 1, \dots N$.

The student and teacher networks share the same architecture design, however, the parameters of the teacher network are an EMA of parameters of the student network, allowing the model to distill knowledge from past iterations of itself. We apply centering and sharpening \cite{caron2021emerging} to the teacher outputs to prevent the framework from collapsing to a degenerate solution. As for the projector architecture $\mathcal{P}$, we follow previous work \cite{zhou2022ibot}. However, we disentangle its head by employing two separate linear layers, one dedicated to CLS token and the other for patch tokens, in order to accommodate for the significantly harder task of recovering tokens corrupted in the adversarial manner. Projected features are then used to meet the patch-level and object-level objectives.

\paragraph{Patch-level objective}
The masked patch modeling (MPM) allows the encoder to extract low-level features from point clouds. It is similar to masked language modeling in BERT \cite{devlin2019bert} or masked image modeling in iBOT \cite{zhou2022ibot} with the main difference of using the auxiliary network to mask inputs. The mask generator outputs occlusions following the procedure explained in Section \ref{sec:mask_generator}. A masked sequence obtained using Equation \eqref{eq:masking}, is passed to the student network which outputs patch tokens projections $\mathbf{\hat{h}}^\text{patch}$. The teacher network processes the original sequence $\mathbf{x}^v$, creating representative patch token projections $\mathbf{h}^\text{patch}$ that we aim to recover under the assumption that each patch token can be recovered independently. In this objective the task is to minimize cross-entropy loss between projected features of the student and the teacher:
\begin{equation}
    \lmpm = -\sum_{i=1}^p \left(\mathbf{h}^\text{patch}_i\right)^{\text{T}} \log \hat{\mathbf{h}}^\text{patch}_i,
\end{equation}
where $\mathbf{h}^\text{patch}_i$ and $\hat{\mathbf{h}}^\text{patch}_i$ are projections for $i$-th patch. The loss is calculated for all of the $N$ masks.

\paragraph{Object-level objective}
In this objective we also consider cross-entropy loss as in MPM objective, however, the self-distillation process is performed on the class token of projected local and global views. Precisely, we minimize:
\begin{equation}
    \lcls = - \left(\mathbf{h}^\text{CLS}\right)^{\text{T}} \log \hat{\mathbf{h}}^\text{CLS},
\end{equation}
where $\mathbf{h}^\text{CLS} = \mathcal{P}_t( \mathcal{F}_t( \mathbf{x}^v ))$ and $\hat{\mathbf{h}}^\text{CLS} = \mathcal{P}_s( \mathcal{F}_s( \mathbf{x}^u ))$. $\lcls$ is optimized mainly between projections of local view and global view but also between projections of different global views. Since the CLS tokens are aggregated representations of the views, optimizing $\lcls$ allows the encoder to learn a global shape features invariant on the augmentations used.

\paragraph{Optimization goals} Finally, the objective for the encoder $\mathcal{P \circ F}$ is to minimize the sum $\lmpm + \lcls$. The mask generator $\mathcal{M}$ aims to minimize:
\begin{equation}
    \label{eq:losses}
    \mathcal{L}_\text{mask} = \alpha \mathcal{L}_\text{spar} + \beta \mathcal{L}_\text{div} - \lmpm - \lcls.
\end{equation}
Note that we change the sign of the encoder's objectives to make it adversarial. For the choice of weighting hyperparameters, please see Section \ref{sec:training_settings}.

%% file: sections/3_experiments.tex
\section{Experiments}
This section presents the pre-training setup and reports results from experiments conducted on various downstream tasks. We also conduct an ablation study to validate the effectiveness of our approach.

\subsection{Self-supervised learning setup}
\label{sec:exp_ssl}
\paragraph{Data} For pre-training we use ShapeNet \cite{chang2015shapenet} dataset which contains $52,472$ unique 3D models from 55 common object categories. We follow the official data split and use training and validation splits for pre-training, excluding the test split. We preprocess each shape in the dataset to prepare uniformly sampled point clouds. First, we sample $16384$ points from provided meshes, and from that we sample $8192$ points using the furthest point sampling (FPS) algorithm. 

During pre-training, we randomly choose $1200$ points and reduce them to $1024$ via FPS. Random scaling, translation, and jitter are applied as data augmentation. We group points into 64 patches with 32 points each following previous work \cite{yu2022pointbert}. We crop 8 local views that cover $10\%$ to $30\%$ of a point cloud and 2 global views that cover $30\%$ to $50\%$. 

\paragraph{Training settings}
\label{sec:training_settings}
Weights of the mask-related loss (Equation \eqref{eq:losses}) are set to $\alpha = 0.2$, $\beta = 0.03$. Network $\mathcal{M}$ is set to generate $N=3$ masks. We conduct the training of \model{} for approximately $33,000$ steps with a batch size of $128$, maintaining a consistent configuration for both the encoder's and mask generator's optimizers, with the exception being the learning rate. Specifically, we employ the AdamW optimizer paired with a cosine schedule with a warmup. The encoder's optimizer is set to a learning rate of $10^{-4}$, while the mask generator's optimizer is set to a learning rate of $3 \times 10^{-5}$. Furthermore, we apply a k-decay \cite{zhang2022kdecay} of $1.5$ and weight decay of $0.05$. Settings for constructing both the encoder $\mathcal{F}$ and the mask generator $\mathcal{M}$ are in Table \ref{tab:settings}. For parameters of the teacher network, we set the initial momentum weight to $0.994$ and anneal it to $1.0$ with a cosine schedule. 

\begin{table}
  \caption{Settings used to construct the encoder $\mathcal{F}$ and the mask generator $\mathcal{M}$ transformers \cite{vaswani2017attention}. }
  \label{tab:settings}
  \centering
  \begin{tabular}{lcc}
    \toprule
    Setting & Encoder $\mathcal{F}$  & Mask Generator $\mathcal{M}$ \\
    \midrule
    depth & 12 & 3 \\
    number of heads & 6 & 4 \\
    stochastic depth & 0.1 & 0.1 \\
    embedding dimension & 384 & 384 \\
    FFN & SwiGLU & SwiGLU \\
    \cmidrule(l{0.5em}r{0.5em}){1-3}
    \#parameters & 21.8M & 6.0M\\
    
    \bottomrule
  \end{tabular}
\end{table}

\subsection{Downstream tasks}
\input{tables/linear_svm}
\paragraph{Linear SVM} We test 3D point cloud representations learned during pretraining by extracting features from the frozen encoder and testing them on ModelNet40 \cite{wu20153d}, a dataset comprising 12,311 synthetic 3D models across 40 categories. We adhere to the official split of $9843$ examples in the training set, with the remainder in the testing set. We report classification accuracy in Table \ref{tab:svm}. \model{} achieves the state-of-the-art result of $91.52\%$ accuracy, representing a  $+1.32$pp improvement among non-hierarchical transformer and generative architectures and second-highest mean accuracy overall. It is also worth noting that our results are averaged and accompanied by error bars, thereby not relying on chance in the commonly reported voting method \cite{liu2019relationshape}. 

\begin{figure}
  \centering
  \includegraphics[width=\linewidth]{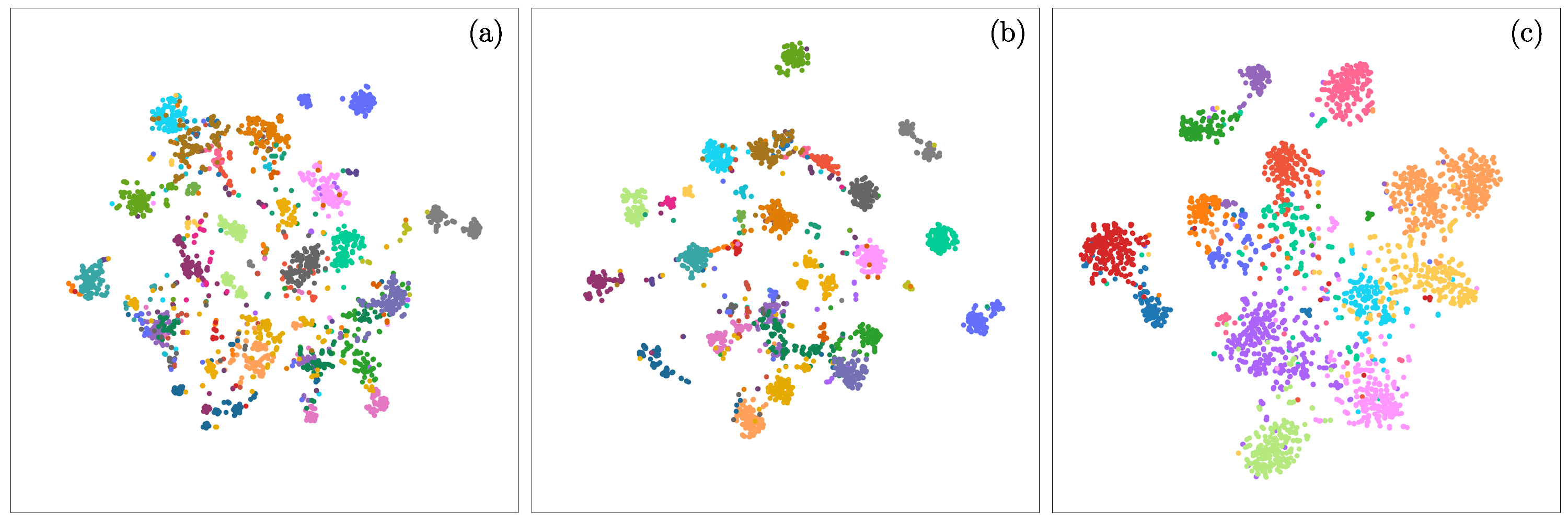}
  \caption{\textbf{t-SNE visualization of learned representations}. We show: (a) features of ModelNet40 extracted from our self-supervised method \model{}, (b) features of ModelNet40 extracted after fine-tuning and (c) features of ScanObjectNN 3D objects after fine-tuning. }
  \label{img:embeddings}
\end{figure}

Additionally, we investigate learned representations using t-SNE projection to 2 components in Figure \ref{img:embeddings}. We show that clusters of different classes from ModelNet40 can be clearly separated even only after our self-supervised pre-training (a). Features after finetuning (b) further improve quality of representations on ModelNet40. We also depict t-SNE projection of features learned during finetuning on the hardest setting of ScanObjectNN dataset (c).

\begin{figure}
  \centering
  \includegraphics[width=0.95\linewidth]{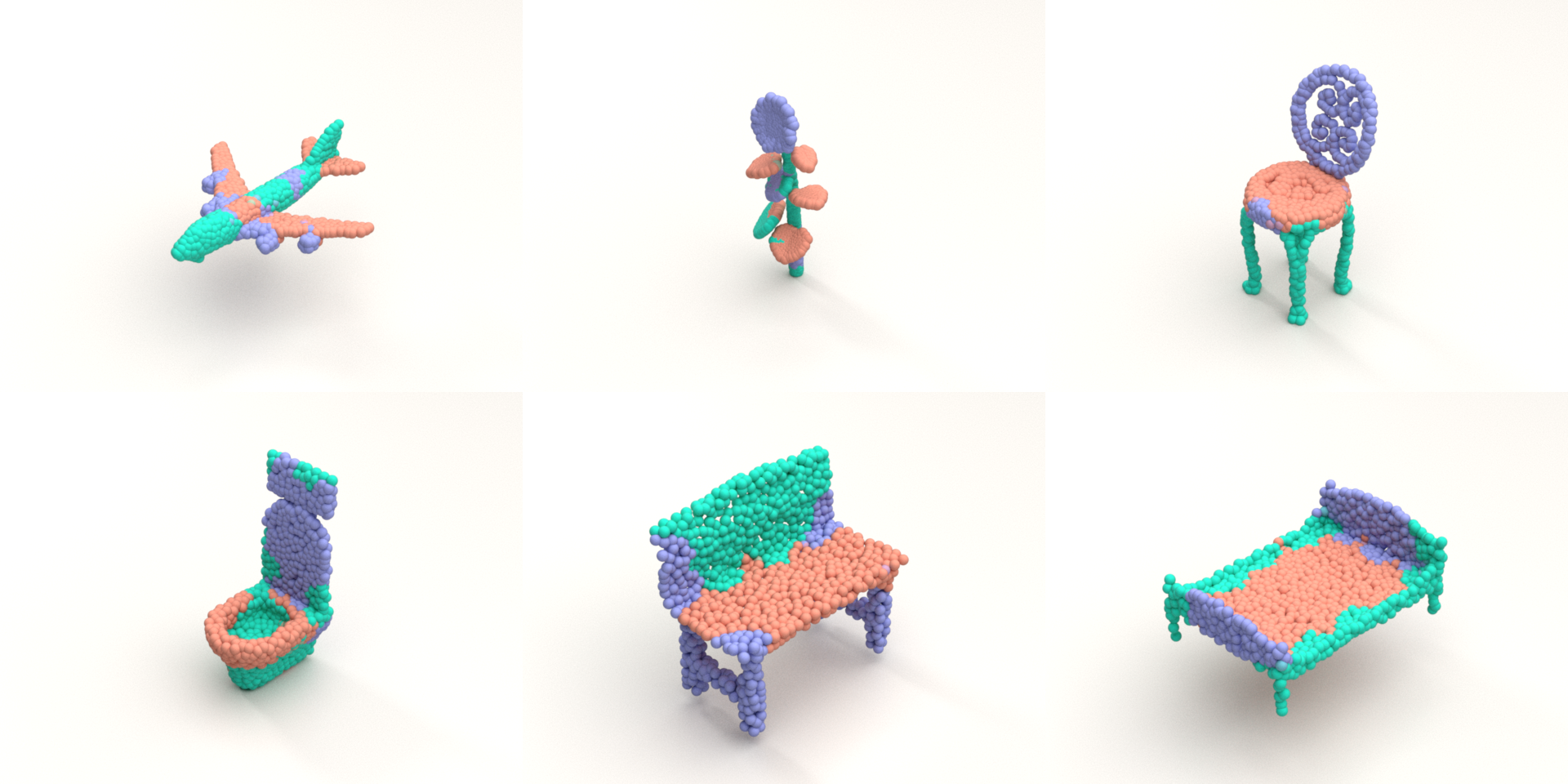}
  \caption{Examples of masks learned by the mask generator $\mathcal{M}$ with $N=3$ during \model{} pre-training. Different colors depict distinct masks.}
  \label{img:mask_renders}
\end{figure}

Furthermore, we provide examples of masks learned by the mask generator $\mathcal{M}$ in Figure \ref{img:mask_renders}. Presented point clouds are divided into $N=3$ masks distinguished by different colors. As we can see, each mask covers semantically meaningful part of a point cloud, fulfilling its desired goal.

\paragraph{Object classification on ModelNet40} We finetune our pretrained model on ModelNet40 with $1024$ points per object. The same data augmentation techniques employed during the pretraining phase are utilized here, specifically random scaling, translation, and jitter. We abstain from using the voting method and instead report the mean accuracy with a standard deviation across three runs in Table \ref{tab:scobjnn_mn40}. Our results are competitive with other methodologies, achieving an accuracy of $92.57\%$.

\input{tables/classification}

\paragraph{Object classification on ScanObjectNN} We finetune our pretrained model on real-world dataset ScanObjectNN \cite{uy2019revisiting}, which comprises $2902$ 3D objects across 15 categories. This dataset presents more challenges than ModelNet40 due to noise factors such as background interference, missing parts, and deformations. In this experiment, we amplify the number of patches per point cloud to $128$, to accommodate an increased number of points of $2048$. Our results are reported across three commonly utilized settings: OBJ-BG, OBJ-ONLY, and PB-T50\_RS. Our model demonstrates state-of-the-art accuracy among non-hierarchical transformer and generative approaches in the OBJ-BG ($90.36\%$) and OBJ-ONLY ($88.35\%$) settings, falling short only to the hierarchical transformer PointM2AE \cite{zhang2022pointm2ae}. Additionally, we report a competitive result in the most challenging setting, PB-T50-RS, achieving an accuracy of $84.49\%$.

\paragraph{Few-shot learning on ModelNet40} We follow previous works \cite{yu2022pointbert, pang2022masked, liu2022masked, zhang2022pointm2ae} in few-shot learning experiment setup. We employ the standard $n$-way, $m$-shot setting, where $n$ is the number of classes and $m$ corresponds to the number of samples per class. 
We provide results for $\{n, m \} \in \{5, 10\} \times \{10, 20\}$ in Table \ref{tab:fewshot} across 10 different folds, consistent with the data splits employed in the referenced works. Our approach achieves notable results, particularly in the 5-way setting, where we outperform the non-hierarchical transformer and generative architectures, achieving improvements of $+0.4$pp and $+0.7$pp in 10-shot and 20-shot settings respectively, while also exhibiting a lower standard deviation. In the more challenging 10-way setting, we also report results that are superior or competitive with the current state-of-the-art.

\input{tables/fewshot}

\paragraph{Part segmentation on ShapeNetPart} We evaluate PointCAM on a challenging part segmentation task that requires the prediction of per-point labels. We use the ShapeNetPart dataset \cite{chang2015shapenet}, which comprises $16,881$ objects with parts across 16 categories. Following the procedure of earlier works \cite{yu2022pointbert, liu2022masked, pang2022masked}, we sample $2048$ points and group them into $128$ patches. We extract features from the 4th, 8th, and 12th layers of the encoder, and employ max pooling in conjunction with global average pooling to generate distinct sets of features. These features are used to upsample the center points of the patches and MLP is applied to predict the class of each point. We report mean Intersection over Union (IoU) across instances ($\text{mIoU}_I$) and categories ($\text{mIoU}_C$) in Table \ref{tab:segmentation}. Our results indicate that PointCAM is competitive with the current state-of-the-art methods. 

\input{tables/segmentation}

\subsection{Ablation study}
The ablation tests were conducted under the same experimental setup as the Linear SVM experiment to ensure a fair comparison. We maintained a consistent model architecture and data, only varying the studied parameter.
To elucidate the benefits of our adversarial masking method, we conducted an ablation study, contrasting it with the conventional approach of random masking.  Our findings, detailed in Table \ref{tab:ablation}, reveal a significant performance advantage when employing our adversarial masking technique, improving the linear SVM evaluation accuracy by $+0.43$pp.
We also investigate the impact of the number of masks generated by $\mathcal{M}$, varying the value of $N$ from 2 to 4. Our results, presented in Table \ref{tab:ablation}, suggest that $N=3$ yields the best representations.

\input{tables/ablation}

%% file: tables/linear_svm.tex
\begin{table}
  \caption{\textbf{Linear SVM evaluation on ModelNet40.} Mean accuracy with a standard deviation from 10 SVM runs with different seeds is reported. Methods using hierarchical or multimodal transformers are in gray creating a separate group for a fair comparison. The best results among the groups are bolded.}
  \label{tab:svm}
  \centering
  \begin{tabular}{lc}
    \toprule
    Method  & Accuracy [\%] \\
    \midrule
    
    3D-GAN \cite{wu2017learning}  & $83.3$  \\
    Latent-GAN \cite{valsesia2018learning} & $85.7$  \\
    SO-Net \cite{li2018sonet}  & $87.3$  \\
    FoldingNet \cite{yang2018foldingnet} & $88.4$  \\
    MAP-VAE \cite{han2019multiangle}  & $88.4$  \\
    VIP-GAN \cite{han2018view} & $90.2$  \\
    PointBERT \cite{yu2022pointbert}  & $87.4$  \\
    \textbf{PointCAM} (ours) & \bm{$91.52 \pm 0.26$}  \\
    \cmidrule(l{0.5em}r{0.5em}){1-2}
    OcCo \cite{wang2021unsupervised}  & {\color[HTML]{9B9B9B} $89.2$}  \\
    CrossPoint \cite{afham2022crosspoint}  & {\color[HTML]{9B9B9B} $91.2$}  \\
    PointM2AE \cite{zhang2022pointm2ae}  & {\color[HTML]{9B9B9B} $\bm{92.9}$}  \\
    
    \bottomrule
  \end{tabular}
\end{table}

%% file: tables/classification.tex
\begin{table}
  \caption{\textbf{Classification on ScanObjectNN and ModelNet40 with fine-tuning.} Mean accuracy with standard deviation from 3 runs with different seeds is reported. We omit voting strategy and report such results for other works if possible. Methods using hierarchical or multimodal transformers are in gray creating a separate group for a fair comparison. The best results among the groups are bolded.}
  \label{tab:scobjnn_mn40}
  \centering
  \begin{tabular}{lcccc}
    \toprule
    \multicolumn{1}{c}{\multirow{2}{*}{Method}} & \multicolumn{3}{c}{ScanObjectNN}  & \multicolumn{1}{c}{ModelNet40} \\
    \cmidrule(l{0.5em}r{0.5em}){2-4}\cmidrule(l{0.5em}r{0.5em}){5-5}
    & OBJ-BG & OBJ-ONLY & PB-T50-RS & 1k points \\
    \midrule
    PointNet \cite{qi2017pointnet}  & $73.3$ & $79.2$ & $68.0$ & $89.2$ \\
    PointNet++ \cite{qi2017pointnetplus}  & $82.3$ & $84.3$ & $77.9$ & $90.7$  \\
    DGCNN \cite{wang2019dynamic}  & $82.8$ & $86.2$ & $78.1$ & \bm{$92.9$} \\
    PointCNN \cite{pointcnn2018}  & $86.1$ & $85.5$ & $78.5$ & $92.2$ \\    
    Transformer \cite{yu2022pointbert}  & $79.86$ & $80.55$ & $77.24$ & $91.4$ \\
    PointBERT \cite{yu2022pointbert}  & $87.43$ & $88.12$ & $83.07$ & $92.7$  \\
    MaskPoint \cite{liu2022masked}  & $89.30$ & $88.10$ & $84.30$ & -- \\
    PointMAE \cite{pang2022masked}  & $90.02$ & $88.29$ & \bm{$85.18$} & $92.7 \pm 0.1$  \\
    \textbf{PointCAM} (ours) & $\bm{90.36 \pm 0.25}$ & $\bm{88.35 \pm 0.53}$ & $84.49 \pm 0.38$ & $92.6 \pm 0.2$  \\
    \cmidrule(l{0.5em}r{0.5em}){1-5}

    PointM2AE \cite{zhang2022pointm2ae}  & {\color[HTML]{9B9B9B}$\textbf{91.22}$} & {\color[HTML]{9B9B9B}$\textbf{88.81}$} & {\color[HTML]{9B9B9B}$\textbf{86.43}$} & {\color[HTML]{9B9B9B}$\textbf{93.4}$}  \\
    
    \bottomrule
  \end{tabular}
\end{table}

%% file: tables/fewshot.tex
\begin{table}
  \caption{\textbf{Few-shot classification on ModelNet40. } Mean accuracy with standard deviation across 10 different folds is reported. Methods using hierarchical or multimodal transformers are in gray creating a separate group for a fair comparison. The best results among the groups are bolded.}
  \label{tab:fewshot}
  \centering
  \begin{tabular}{lcccc}
    \toprule
    \multicolumn{1}{c}{\multirow{2}{*}{Method}}  & \multicolumn{2}{c}{5-way} & \multicolumn{2}{c}{10-way}  \\
    \cmidrule(l{0.5em}r{0.5em}){2-3}
    \cmidrule(l{0.5em}r{0.5em}){4-5}
    & 10-shot & 20-shot & 10-shot & 20-shot \\
    \midrule

    DGCNN \cite{wang2019dynamic} & $91.8 \pm 3.7$ & $93.4 \pm 3.2$ & $86.3 \pm 6.2$ & $90.9 \pm 5.1$  \\    
    OcCo \cite{wang2021unsupervised} & $94.0 \pm 3.6$ & $95.9 \pm 2.3$ & $89.4 \pm 5.1$ & $92.4 \pm 4.6$  \\
    Transformer \cite{yu2022pointbert}  & $87.8 \pm 5.2$ & $93.3 \pm 4.3$ & $84.6 \pm 5.5$ & $89.4 \pm 6.3$  \\
    PointBERT \cite{yu2022pointbert}  & $94.6 \pm 3.1$ & $96.3 \pm 2.7$ & $91.0 \pm 5.4$ & $92.7 \pm 5.1$  \\
    MaskPoint \cite{liu2022masked}  & $95.0 \pm 3.7$ & $97.2 \pm 1.7$ & $92.6 \pm 4.1$ & \bm{$95.0 \pm 3.0$}  \\
    PointMAE \cite{pang2022masked} & $96.3 \pm 2.5$ & $97.8 \pm 1.8$ & $92.6 \pm 4.1$ & \bm{$95.0 \pm 3.0$}  \\
    \textbf{PointCAM} (ours) & \bm{$96.7 \pm 2.1$} & \bm{$98.5 \pm 1.6$} & \bm{$92.8 \pm 4.2$} & $94.9 \pm 3.4$  \\
    \cmidrule(l{0.5em}r{0.5em}){1-5}

    PointM2AE \cite{zhang2022pointm2ae}  & \color[HTML]{9B9B9B}\bm{$96.8 \pm 1.8$} & \color[HTML]{9B9B9B} \bm{$98.9 \pm 1.2$} & \color[HTML]{9B9B9B} \bm{$93.3 \pm 3.9$} & \color[HTML]{9B9B9B}\bm{ $95.8 \pm 3.0$}   \\
    \bottomrule
  \end{tabular}
\end{table}

%% file: tables/segmentation.tex
\begin{table}
  \caption{\textbf{Part segmentation on ShapeNetPart.} Mean IoU across part categories $\text{mIoU}_C$ and mean IoU across instances $\text{mIoU}_I$ is reported. Methods using hierarchical or multimodal transformers are in gray creating a separate group for a fair comparison. The best results among the groups are bolded.}
  \label{tab:segmentation}
  \centering
  \begin{tabular}{lcc}
    \toprule
    Method & $\text{mIoU}_C$ &  $\text{mIoU}_I$ \\
    \midrule
    
    PointNet \cite{qi2017pointnet} & $80.4$ & $83.7$ \\
    PointNet++ \cite{qi2017pointnetplus}  & $81.9$ & $85.1$  \\
    DGCNN \cite{wang2019dynamic} & $82.3$ & $85.2$ \\    
    Transformer \cite{yu2022pointbert} & $83.4$  & $85.1$ \\
    PointBERT \cite{yu2022pointbert} & $84.1$  & $85.6$ \\
    MaskPoint \cite{liu2022masked}  & \bm{$84.4$} & $86.0$  \\
    PointMAE \cite{pang2022masked} & $84.2$ & \bm{$86.1$} \\
    \textbf{PointCAM} (ours) & $84.1$ & $86.0$ \\
    \cmidrule(l{0.5em}r{0.5em}){1-3}
    PointM2AE \cite{zhang2022pointm2ae}  & \color[HTML]{9B9B9B}\bm{$84.9$} & \color[HTML]{9B9B9B}\bm{$86.5$}   \\
    \bottomrule
  \end{tabular}
\end{table}

%% file: tables/ablation.tex
\begin{table}
  \caption{\textbf{Ablation study.} Different masking strategies are evaluated with linear SVM. Settings are the same as in Linear SVM experiment, only masking strategy varies. Mean accuracy with standard deviation of 10 runs from different seeds is reported.}
  \label{tab:ablation}
  \centering
  \begin{tabular}{lccc}
    \toprule
    Masking strategy & Number of masks & Masking ratio & Accuracy [\%] \\
    \midrule
    adversarial (ours) & 4 & - & $90.99 \pm 0.24$ \\ 
    adversarial (ours) & 3 & - & \bm{$91.52 \pm 0.26$} \\
    adversarial (ours) & 2 & - & $91.17 \pm 0.19$ \\ 
    random & - & 0.10--0.45 & $91.09 \pm 0.32$ \\ 
    block & - & 0.10--0.45 & $91.01 \pm 0.30$ \\ 
    \bottomrule
  \end{tabular}
\end{table}

%% file: sections/4_conclusion.tex
\section{Conclusion}

In this work, we introduce \model{}, a novel adversarial masking method for self-supervised learning with 3D point cloud data. Our approach diverges from the conventional random masking strategies, instead utilizing a network to learn an optimal masking function. Our extensive experiments demonstrate that \model{} can effectively enhance the learning capability of transformer models. It achieves state-of-the-art or competitive performance among methods based on transformer architectures on several benchmarks, including classification, segmentation, and few-shot learning tasks. Importantly, our ablation studies highlight the significant contribution of our adversarial masking method to these performance gains.

\paragraph{Limitations} While the results are promising, PointCAM's performance falls short when compared to more sophisticated architectures, such as hierarchical transformers. We are also aware that optimizing sparsity penalty $\mathcal{L}_\text{spar}$ leads to generating masks of equal size, which is not always desired since semantically meaningful parts of 3D objects can have various sizes. Furthermore, by applying these masks to patches, instead of points, we decrease the resolution of masks, making lines between them deficient.

%% file: ms.bbl
\begin{thebibliography}{10}

\bibitem{achlioptas2018learning}
Panos Achlioptas, Olga Diamanti, Ioannis Mitliagkas, and Leonidas Guibas.
\newblock Learning representations and generative models for 3d point clouds, 2018.

\bibitem{afham2022crosspoint}
Mohamed Afham, Isuru Dissanayake, Dinithi Dissanayake, Amaya Dharmasiri, Kanchana Thilakarathna, and Ranga Rodrigo.
\newblock Crosspoint: Self-supervised cross-modal contrastive learning for 3d point cloud understanding, 2022.

\bibitem{baevski2022data2vec}
Alexei Baevski, Wei-Ning Hsu, Qiantong Xu, Arun Babu, Jiatao Gu, and Michael Auli.
\newblock data2vec: A general framework for self-supervised learning in speech, vision and language, 2022.

\bibitem{caron2020unsupervised}
Mathilde Caron, Ishan Misra, Julien Mairal, Priya Goyal, Piotr Bojanowski, and Armand Joulin.
\newblock Unsupervised learning of visual features by contrasting cluster assignments.
\newblock {\em Advances in neural information processing systems}, 33:9912--9924, 2020.

\bibitem{caron2021emerging}
Mathilde Caron, Hugo Touvron, Ishan Misra, Hervé Jégou, Julien Mairal, Piotr Bojanowski, and Armand Joulin.
\newblock Emerging properties in self-supervised vision transformers, 2021.

\bibitem{chang2015shapenet}
Angel~X. Chang, Thomas Funkhouser, Leonidas Guibas, Pat Hanrahan, Qixing Huang, Zimo Li, Silvio Savarese, Manolis Savva, Shuran Song, Hao Su, Jianxiong Xiao, Li~Yi, and Fisher Yu.
\newblock Shapenet: An information-rich 3d model repository, 2015.

\bibitem{chen2020simple}
Ting Chen, Simon Kornblith, Mohammad Norouzi, and Geoffrey Hinton.
\newblock A simple framework for contrastive learning of visual representations.
\newblock In {\em International conference on machine learning}, pages 1597--1607. PMLR, 2020.

\bibitem{chorowski2021aligned}
Jan Chorowski, Grzegorz Ciesielski, Jaros{\l}aw Dzikowski, Adrian {\L}a{\'n}cucki, Ricard Marxer, Mateusz Opala, Piotr Pusz, Pawe{\l} Rychlikowski, and Micha{\l} Stypu{\l}kowski.
\newblock Aligned contrastive predictive coding.
\newblock {\em arXiv preprint arXiv:2104.11946}, 2021.

\bibitem{devlin2019bert}
Jacob Devlin, Ming-Wei Chang, Kenton Lee, and Kristina Toutanova.
\newblock Bert: Pre-training of deep bidirectional transformers for language understanding, 2019.

\bibitem{han2018view}
Zhizhong Han, Mingyang Shang, Yu-Shen Liu, and Matthias Zwicker.
\newblock View inter-prediction gan: Unsupervised representation learning for 3d shapes by learning global shape memories to support local view predictions, 2018.

\bibitem{han2019multiangle}
Zhizhong Han, Xiyang Wang, Yu-Shen Liu, and Matthias Zwicker.
\newblock Multi-angle point cloud-vae: Unsupervised feature learning for 3d point clouds from multiple angles by joint self-reconstruction and half-to-half prediction, 2019.

\bibitem{he2022masked}
Kaiming He, Xinlei Chen, Saining Xie, Yanghao Li, Piotr Doll{\'a}r, and Ross Girshick.
\newblock Masked autoencoders are scalable vision learners.
\newblock In {\em Proceedings of the IEEE/CVF Conference on Computer Vision and Pattern Recognition}, pages 16000--16009, 2022.

\bibitem{he2020momentum}
Kaiming He, Haoqi Fan, Yuxin Wu, Saining Xie, and Ross Girshick.
\newblock Momentum contrast for unsupervised visual representation learning.
\newblock In {\em Proceedings of the IEEE/CVF conference on computer vision and pattern recognition}, pages 9729--9738, 2020.

\bibitem{hendrycks2020gaussian}
Dan Hendrycks and Kevin Gimpel.
\newblock Gaussian error linear units (gelus), 2020.

\bibitem{li2018sonet}
Jiaxin Li, Ben~M. Chen, and Gim~Hee Lee.
\newblock So-net: Self-organizing network for point cloud analysis, 2018.

\bibitem{pointcnn2018}
Yangyan Li, Rui Bu, Mingchao Sun, Wei Wu, Xinhan Di, and Baoquan Chen.
\newblock Pointcnn: Convolution on x-transformed points.
\newblock In S.~Bengio, H.~Wallach, H.~Larochelle, K.~Grauman, N.~Cesa-Bianchi, and R.~Garnett, editors, {\em Advances in Neural Information Processing Systems}, volume~31. Curran Associates, Inc., 2018.

\bibitem{liu2022masked}
Haotian Liu, Mu~Cai, and Yong~Jae Lee.
\newblock Masked discrimination for self-supervised learning on point clouds, 2022.

\bibitem{liu2019relationshape}
Yongcheng Liu, Bin Fan, Shiming Xiang, and Chunhong Pan.
\newblock Relation-shape convolutional neural network for point cloud analysis, 2019.

\bibitem{pang2022masked}
Yatian Pang, Wenxiao Wang, Francis E.~H. Tay, Wei Liu, Yonghong Tian, and Li~Yuan.
\newblock Masked autoencoders for point cloud self-supervised learning, 2022.

\bibitem{qi2017pointnet}
Charles~R. Qi, Hao Su, Kaichun Mo, and Leonidas~J. Guibas.
\newblock Pointnet: Deep learning on point sets for 3d classification and segmentation, 2017.

\bibitem{qi2017pointnetplus}
Charles~R. Qi, Li~Yi, Hao Su, and Leonidas~J. Guibas.
\newblock Pointnet++: Deep hierarchical feature learning on point sets in a metric space, 2017.

\bibitem{sauder2019selfsupervised}
Jonathan Sauder and Bjarne Sievers.
\newblock Self-supervised deep learning on point clouds by reconstructing space, 2019.

\bibitem{shazeer2020glu}
Noam Shazeer.
\newblock Glu variants improve transformer, 2020.

\bibitem{shi2022adversarial}
Yuge Shi, N.~Siddharth, Philip H.~S. Torr, and Adam~R. Kosiorek.
\newblock Adversarial masking for self-supervised learning, 2022.

\bibitem{uy2019revisiting}
Mikaela~Angelina Uy, Quang-Hieu Pham, Binh-Son Hua, Duc~Thanh Nguyen, and Sai-Kit Yeung.
\newblock Revisiting point cloud classification: A new benchmark dataset and classification model on real-world data, 2019.

\bibitem{valsesia2018learning}
Diego Valsesia, Giulia Fracastoro, and Enrico Magli.
\newblock Learning localized generative models for 3d point clouds via graph convolution.
\newblock In {\em International Conference on Learning Representations}, 2019.

\bibitem{van2020scan}
Wouter Van~Gansbeke, Simon Vandenhende, Stamatios Georgoulis, Marc Proesmans, and Luc Van~Gool.
\newblock Scan: Learning to classify images without labels.
\newblock In {\em Computer Vision--ECCV 2020: 16th European Conference, Glasgow, UK, August 23--28, 2020, Proceedings, Part X}, pages 268--285. Springer, 2020.

\bibitem{vaswani2017attention}
Ashish Vaswani, Noam Shazeer, Niki Parmar, Jakob Uszkoreit, Llion Jones, Aidan~N. Gomez, Lukasz Kaiser, and Illia Polosukhin.
\newblock Attention is all you need, 2017.

\bibitem{wang2021unsupervised}
Hanchen Wang, Qi~Liu, Xiangyu Yue, Joan Lasenby, and Matthew~J. Kusner.
\newblock Unsupervised point cloud pre-training via occlusion completion, 2021.

\bibitem{wang2019dynamic}
Yue Wang, Yongbin Sun, Ziwei Liu, Sanjay~E. Sarma, Michael~M. Bronstein, and Justin~M. Solomon.
\newblock Dynamic graph cnn for learning on point clouds, 2019.

\bibitem{wu2017learning}
Jiajun Wu, Chengkai Zhang, Tianfan Xue, William~T. Freeman, and Joshua~B. Tenenbaum.
\newblock Learning a probabilistic latent space of object shapes via 3d generative-adversarial modeling, 2017.

\bibitem{wu20153d}
Zhirong Wu, Shuran Song, Aditya Khosla, Fisher Yu, Linguang Zhang, Xiaoou Tang, and Jianxiong Xiao.
\newblock 3d shapenets: A deep representation for volumetric shapes, 2015.

\bibitem{xie2020pointcontrast}
Saining Xie, Jiatao Gu, Demi Guo, Charles~R. Qi, Leonidas~J. Guibas, and Or~Litany.
\newblock Pointcontrast: Unsupervised pre-training for 3d point cloud understanding, 2020.

\bibitem{yan2023implicit}
Siming Yan, Zhenpei Yang, Haoxiang Li, Chen Song, Li~Guan, Hao Kang, Gang Hua, and Qixing Huang.
\newblock Implicit autoencoder for point cloud self-supervised representation learning, 2023.

\bibitem{yang2018foldingnet}
Yaoqing Yang, Chen Feng, Yiru Shen, and Dong Tian.
\newblock Foldingnet: Point cloud auto-encoder via deep grid deformation, 2018.

\bibitem{yu2022pointbert}
Xumin Yu, Lulu Tang, Yongming Rao, Tiejun Huang, Jie Zhou, and Jiwen Lu.
\newblock Point-bert: Pre-training 3d point cloud transformers with masked point modeling, 2022.

\bibitem{zamorski2020adversarial}
Maciej Zamorski, Maciej Zi{\k{e}}ba, Piotr Klukowski, Rafa{\l} Nowak, Karol Kurach, Wojciech Stokowiec, and Tomasz Trzci{\'n}ski.
\newblock Adversarial autoencoders for compact representations of 3d point clouds.
\newblock {\em Computer Vision and Image Understanding}, 193:102921, 2020.

\bibitem{zhang2022pointm2ae}
Renrui Zhang, Ziyu Guo, Rongyao Fang, Bin Zhao, Dong Wang, Yu~Qiao, Hongsheng Li, and Peng Gao.
\newblock Point-m2ae: Multi-scale masked autoencoders for hierarchical point cloud pre-training, 2022.

\bibitem{zhang2022kdecay}
Tao Zhang and Wei Li.
\newblock kdecay: Just adding k-decay items on learning-rate schedule to improve neural networks, 2022.

\bibitem{zhang2021selfsupervised}
Zaiwei Zhang, Rohit Girdhar, Armand Joulin, and Ishan Misra.
\newblock Self-supervised pretraining of 3d features on any point-cloud, 2021.

\bibitem{zheng2019learning}
Heliang Zheng, Jianlong Fu, Zheng-Jun Zha, and Jiebo Luo.
\newblock Learning deep bilinear transformation for fine-grained image representation, 2019.

\bibitem{zhou2022ibot}
Jinghao Zhou, Chen Wei, Huiyu Wang, Wei Shen, Cihang Xie, Alan Yuille, and Tao Kong.
\newblock ibot: Image bert pre-training with online tokenizer, 2022.

\bibitem{zhou20223doae}
Junsheng Zhou, Xin Wen, Baorui Ma, Yu-Shen Liu, Yue Gao, Yi~Fang, and Zhizhong Han.
\newblock 3d-oae: Occlusion auto-encoders for self-supervised learning on point clouds, 2022.

\end{thebibliography}
